\pdfoutput=1
\documentclass[letterpaper]{article} % DO NOT CHANGE THIS
\usepackage{aaai23}  % DO NOT CHANGE THIS
\usepackage{times}  % DO NOT CHANGE THIS
\usepackage{helvet}  % DO NOT CHANGE THIS
\usepackage{courier}  % DO NOT CHANGE THIS
\usepackage[hyphens]{url}  % DO NOT CHANGE THIS
\usepackage{graphicx} % DO NOT CHANGE THIS
\usepackage{natbib}  % DO NOT CHANGE THIS AND DO NOT ADD ANY OPTIONS TO IT
\usepackage{caption} % DO NOT CHANGE THIS AND DO NOT ADD ANY OPTIONS TO IT
\usepackage{algorithm}
\usepackage{algorithmic}
\usepackage{amsmath}

\usepackage{newfloat}
\usepackage{listings}
\DeclareCaptionStyle{ruled}{labelfont=normalfont,labelsep=colon,strut=off} % DO NOT CHANGE THIS
\floatstyle{ruled}
\newfloat{listing}{tb}{lst}{}
\floatname{listing}{Listing}
\nocopyright 
\title{Towards A Natural Language Interface \\ for Flexible Multi-Agent Task Assignment}
\author{
    Jake Brawer\textsuperscript{\rm 1 2},
    Kayleigh Bishop\textsuperscript{\rm 1},
    Bradley Hayes\textsuperscript{\rm 1},
    Alessandro Roncone\textsuperscript{\rm 1}
}
\title{Towards A Natural Language Interface \\ for Flexible Multi-Agent Task Assignment}
\author {
       Jake Brawer\textsuperscript{\rm 1 2},
    Kayleigh Bishop\textsuperscript{\rm 1},
    Bradley Hayes\textsuperscript{\rm 1},
    Alessandro Roncone\textsuperscript{\rm 1}
}
\affiliations {
    \textsuperscript{\rm 1} Department of Computer Science, University of Colorado Boulder, 1111 Engineering Drive, Boulder, CO USA\\
    \textsuperscript{\rm 2}  US Army Research Laboratory, Aberdeen Proving Ground, MD 21005, USA\\
   firstname.lastname@colorado.edu
}

\usepackage{bibentry}
\begin{document}

\maketitle
%Here's the grant number you should add W911NF-21-2-02905 (please check past pubs for how we write this part). 
%Figure 1 should be two columns, Figure 2 should be one column. Also, I would beef Fig 1 up.
%You should add more equations! You have the equations in your google slide presentation, just add them here.

\begin{figure*}
    \begin{center}
        \includegraphics[width=\textwidth]{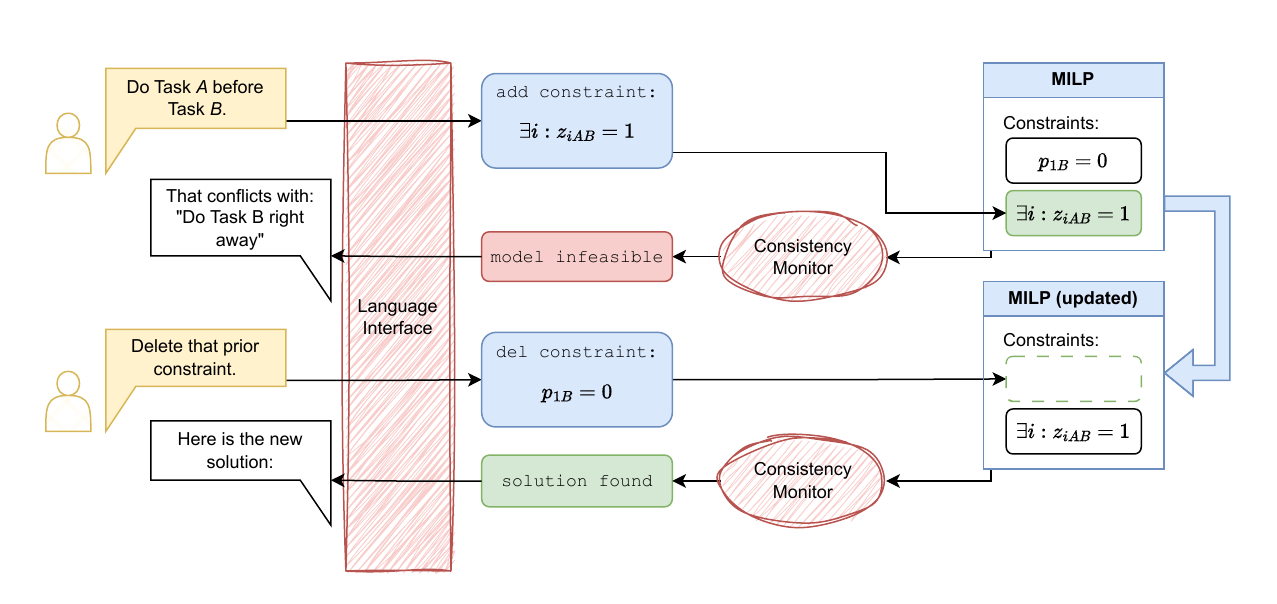}
    \end{center}
    \caption{An example interaction with our proposed system. Our approach will enable users to modify and inspect a multi-agent task assignment system via the interface of spoken language. This will be accomplished by converting user commands into mathematical constraints on an associated MILP using an LLM. The consistency monitor ensures that user-provided constraints do not prevent task assignments from being found, and engages the user in a corrective dialogue if problematic constraints are identified.}\label{fig:overview}
\end{figure*}

\begin{abstract}
Task assignment and scheduling algorithms are powerful tools for autonomously coordinating large teams of robotic or AI agents. However, the decisions these system make often rely on components designed by domain experts, which can be difficult for non-technical end-users to understand or modify to their own ends. In this paper we propose a preliminary design for a flexible natural language interface for a task assignment system. 
The goal of our approach is both to grant  users more control over a task assignment system's decision process, as well as render these decisions more transparent. Users can direct the task assignment system via natural language commands, which are applied as constraints to a mixed-integer linear program (MILP) using a large language model (LLM). Additionally, our proposed system can alert users to potential issues with their commands, and engage them in a corrective dialogue in order to find a viable solution. We conclude with a description of our planned user-evaluation in the simulated environment \textit{Overcooked} and describe next steps towards developing a flexible and transparent task allocation system. 

\end{abstract}

\section{Introduction}

In this paper we propose a flexible natural language interface for a task assignment and scheduling system. 
Task assignment and scheduling algorithms are powerful tools for coordinating large teams of robots, AIs, or humans while demanding minimal user oversight. As demonstrated by \citet{gombolay2015decision}, such systems are not only generally superior to users in delegating tasks but that users prefer to cede control to them.
%Indeed, it has been shown \cite{gombolay2015decision} that such systems are not only generally better than users at delegating tasks, but that users prefer to cede control to these systems. 
However, this finding has been challenged by work demonstrating a reversal in this preference when a user's workload is low \cite{karakikes2023effect}, such as when users are acting in a supervisory role.
Indeed, when empowered to provide supervisory guidance to task assignment systems, users can sometimes quickly infer crucial parts of the optimal solution that would otherwise require substantial time to compute \cite{petersen2013fast}. 
%A further complication is that the task assignment problem is often posed as a  constrained mixed-integer linear program (MILP; \citealp{gombolay2013fast, omar2019probabilistic, wang2020learning}), which can be difficult for non-technical users to modify, let alone understand. 
A further complication is that the task assignment problem is often posed as a  constrained mixed-integer linear program (MILP; \citealp{gombolay2013fast, omar2019probabilistic, wang2020learning}), which can be difficult to modify, let alone understand, without technical expertise or the use of inflexible, task-dependent GUIs.  
% Ideally, we can leverage the power of such algorithms while granting users more control over and, insight into, how task assignments are being made. 
We aim to combine the power of algorithmic approaches with the intuition and experience of human supervisors by giving users the ability to easily control and modify task scheduling formulations.

Our work draws upon research interrogating the use of natural language interfaces for task assignment and scheduling, robot policy learning and motion planning, as well as recent innovations in transformer-based large language models (LLMS; \citealp{vaswani2017attention}).
Work in the former has demonstrated that user-provided symbolic  constraints can be used to effectively guide a wide range of decision making and learning processes, including policy shaping \cite{brawer2023interactive} and  multi-agent task assignment \cite{petersen2013fast}.
In terms of the latter, recent work has shown that LLMs can both transform unrestricted natural language prompts into actionable representations, such as robot policy code \cite{liang2023code, huang2023instruct2act}, as well as produce concise and accurate text summaries of extant source code (e.g. \citealp{ahmed2022few}).
% robot policy learning and execution \cite{van2022correct,brawer2023interactive}. }}
%, realized as symbolic rules applied to a robot's policy \cite{van2022correct,brawer2023interactive}  or as cost functions \cite{sharma2022correcting} but grounded in natural language commands, can produce immediate changes to a robot's policy or plan in a user-friendly manner.
Our approach combines both strands of research by allowing users to influence the task allocation process through natural language directives and using LLMs to enforce these directives as valid constraints on an underlying MILP. Crucially, our system can alert users when added constraints conflict with previously specified constraints, or prevent a satisficing solution from being found, engaging users in a corrective dialogue to amend or remove the faulty constraint.
Additionally, users can interrogate the system as to why certain assignments were made, improving the transparency of the entire process.

In the remainder of this paper, we provide an overview of the proposed approach, as well as outline a planned user-study for evaluating the efficacy and ease-of-use of the approach in a collaborative gridworld environment based on the game \textit{Overcooked} \cite{carroll2019utility}.

\section{Method}
In this section we detail our proposed task-assignment dialogue interface. 
We first define the task-assignment problem for multiple robots moving and working in the same physical space as a mixed-integer linear program (MILP).
Subsequently, we outline proposed architectural components that enable a naive user-in-the-loop to modify and query the task assignment system.
Refer to Fig. \ref{fig:overview} for a high-level diagram of our approach.

%\subsection{Problem Formulation}
%\begin{figure}[t]
%\centering
%\includegraphics[width=\columnwidth]{gfx/diagram.png} % Reduce the figure size so that it is slightly narrower than the column. Don't use precise values for figure width.This setup will avoid overfull boxes.
%\caption{Overview of our proposed system. Our approach will enable users to modify and inspect a multi-agent task assignment system via the interface of spoken language. This will be accomplished by converting user commands into mathematical constraints on an associated MILP (green box) using an LLM (blue box). The consistency monitor (red box) ensures that user-provided constraints do not prevent task assignments from being found, and engages the user in a corrective dialogue if problematic constraints are identified.}
%\label{fig:diagram}
%\end{figure}

Our approached builds upon and extends the method outlined by \citet{petersen2013fast}. Here the authors propose a system enabling users to add constraints to a task-assignment and scheduling MILP designed to coordinate teams of robots. Our work builds on this in two important ways: 1) constraints will be specified entirely via natural language, rather than a bespoke, task-dependent GUI interface; 2) our system is specifically designed to be robust to the challenges and ambiguities inherent to natural language interactions as well as the complexities of real-time system adaptation made possible by such interactions.  To do this we adopt the same basic formalism developed by \citet{petersen2013fast}:
%Our proposed approach uses a modified version of the Tersio algorithm \cite{gombolay2013fast}, a fast scheduling problem for mult-agent tasks.
%As Tersio can find satisficing solutions in real-time to multi-agent tasks with spatio-temporal constraints, it is ideal for the proposed domain where tasks may need to be constantly recalculated due to the influx of user provided constraints. A simplified mathematical formulation of our constrained MILP is provided below:

\begin{align}
& \text{minimize}    && \sum_{i \in R}\sum_{j_{1} \in T \cup \{0\}}\sum_{j_{2} \in T} z_{ij_{1}j_{2}}\cdot k_{ij_{1}j_{2}} \label{eq1}\\
& && - \sum_{i \in R} \sum_{j \in J} x_{ij}\cdot\rho_{j}  + \sum_{j\in T}\left(p_{1j}\cdot\alpha_{1} + p_{2j}\cdot\alpha_{2}\right) \nonumber\\
& \text{subject to} \nonumber \\
& && C_{pre}(A,S, P) \nonumber \\
& && C_{user}(A,S, P |  U) \nonumber\\\nonumber
\end{align}

Here  $x_{ij}, z_{ij_{1}j_{2}} \in A$ for agents $i \in R$ and tasks $j, j_{1}, j_{2} \in T$ are binary decision variables indicating the assignment of an agent $i$ to task $j$ and the order in which tasks are assigned, respectively. $k_{ij_{1}j_{2}}, \rho_{j} \in P$ are scalar parameters encoding the cost incurred by agent $i$ for attempting task $j_{2}$ after $j_{1}$, and the reward for completing task $j$, respectively. $p_{1j}, p_{2j} \in S$ are continuous decision variables weighted by scalar parameters $\alpha_{1},\alpha_{2} \in P$ reflecting how early or late a task $j$ has been scheduled, respectively. 
In sum, this objective seeks to find assignments and schedules that maximize the reward accrued while minimizing the costs incurred by these assignments and the deviations of task schedules from their pre-specified time windows.

%In particular, they indicate  is a list of pre-specified parameters that will shape the solutions found by the MILP including the sets $R$ of agents, $J$ of tasks, and coefficients like $\rho_{j}, \forall j \in J$ encoding the reward for completing task $j$.  $A$ contains sets of binary decision variables controlling task assignments, including the set $X$ where each element $x_{i,j} \in X$ is 1 if agent $i \in R$ is assigned to task $j \in J$, and 0 otherwise. $S$ is contains sets of binary and continuous decision variables responsible for scheduling agents and tasks, such as $t_{i}^{S}$ and $t_{i}^{E}$, which represent the start and end time, respectively, of each task $j \in J$.

$C_{pre}$ and $C_{user}$ are sets of constraints applied to the MILP. $C_{pre}$ is a minimal set of pre-specified constraints comprised of the decision variables and parameters in $A, S$ and $P$. These constraints ensure that a baseline, useful solution can be found for a wide range of multi-agent coordination problems. For example, the constraint $c_{n} \in C_{pre}$ where $c_{n} = \sum_{i \in R} x_{ij} \leq 1, \forall j \in T$ ensures that only one agent $i \in R$ is assigned to one task $j \in T$ at a time. 

$C_{user}$ is a set of user-defined constraints also comprised of the symbols in $A, S$ and $P$, but conditional on user utterances $U$ issued to the language interface (see following section). These constraints encode a user's high-level preferences or the assignment and scheduling process. For example, if a user identifies that agents $a$ and $b$ collaborate well on a task $j$, a user might tell the language interface ``assign agent $a$ to task $i$ if agent $b$ has already been assigned to it," which would be translated to the corresponding conditional hard constraint $x_{aj} \geq x_{bj}$.

\subsection{Language Interface }

%We intend the language interface to act as a bidirectional interface between the task assignment process and the users preferences for the assignment process; it will allow users to both ``program" the system by issuing verbal commands that will be converted into constraints applied to the MILP, as well as ``debug" it, by engaging the user in corrective dialogue when issues arise. Programming of these constraints will be enabled by a speech-to-text system connected to a prompt-engineered LLM. As MILP constraints are  mathematical expressions, the LLM will be prompted to produce responses in terms of the variables contained in $A, S,$ and $P$.
We intend the language interface to act as a bidirectional interface between the task assignment process and the user's preferences for the assignment process; it will allow users to both ``program" the system by issuing verbal commands that will be converted into constraints applied to the MILP, as well as ``debug" it, by engaging the user in corrective dialogue when issues arise.  Ultimately this requires of our language interface two key abilities: i) the ability to transform language into a corresponding mathematical constraint in the form of a syntactically correct MILP optimization API call (Gurobi in this instance; \citealp{pedroso2011optimization}), and ii) the ability to transform these same API calls back into comprehensible text explanations.  To do this we will adapt recent methods for eliciting accurate task-oriented dialogue from LLMs \cite{hudevcek2023llms, yu2023language}. The main idea is to dynamically construct prompts, containing few-shot examples, based both on the system and task state as well as the user utterance itself. This is achieved in part by having the LLM perform multiple passes over a user utterance. In the initial, pre-processing pass, the LLM first attempts to discern the action the user wishes the system to perform (e.g. to add or remove a constraint, to produce an explanation, etc.). This determination is then used to select a more targeted prompt and set of examples for eliciting an accurate and relevant response from the LLM given the same user utterance.
%This is achieved in part by initially prompting the LLM to perform a pre-processing step on the user utterance in order to determine what type of response (e.g. API call or explanation) is required. This determination is then used to to select the appropriate for eliciting a more accurate and relevant action from the LLM.
%Programming of these constraints will be enabled by a speech-to-text system connected to a prompt-engineered LLM. As MILP constraints are  mathematical expressions, the LLM will be prompted to produce responses in terms of the variables contained in $A, S,$ and $P$.

As a consequence, the language interface can enable a user to debug ``compile-time" and ``run-time" errors raised by the task assignment system, to extend the programming metaphor further. Compile-time errors refer here to a failure of the system to find a valid solution to the MILP. Such a failure might be due to a user issuing commands that produce overly restrictive constraints, or constraints that are in conflict. As outlined in the following section, in order to correct this type of error, the language interface would engage the user in a corrective dialogue in order to remove or amend the offending constraints. %This would occur when the language interface failed to map the user utterance to the correct constraint

Run-time errors in the assignment process can occur if tasks are assigned, but do not line up with the user's communicated preferences. In this case, the user would have to take a more proactive role in the debugging process. The language interface will accommodate this by allowing users to issue interrogative, ``why?" queries, such as, ``why wasn't agent $i$ assigned to task $j$?" Such a query would initiate a \textit{counterfactual assignment} wherein the consistency monitor would attempt to make this assignment, and observe which constraints, if any, are violated. The violated constraints are then sent to the language interface, and a suitable answer to the user's query is generated.

%regarding particular task assignments. That is, the user can ask the language interface why certain task assignments were made, and the interface can produce a response using the counterfactual assignment process described in the following section. 

\subsection{Consistency Monitor}

As we are designing our system with the intention that users will be regularly adding to and modifying its constraints, it's imperative that it be able to quickly check the consistency of these constraints.That is, it is conceivable that a user may inadvertently ``break" the task assignment, either by suggesting two constraints that are in direct conflict (e.g. the constraints ``task $a$ should be completed after task $b$" followed by the constraint ``task $b$ should be completed after task $a$"), or a constraint that is overly restrictive. 

In order to mitigate these potential challenges, our system will utilize a three-phased consistency check. In the first phase, our system will perform a \textit{semantic check}, wherein we leverage an LLM's implicit common-sense and linguistic knowledge to check if any of the issued natural language commands are in conflict. After a user has issued a command via the language interface, the consistency monitor prompts the language interface's LLM to check this command against commands that have been previously issued. If a potential conflict is identified, the language interface will communicate this to the user and engage them in a corrective dialogue. During this dialogue the language interface will prompt the user to either amend or remove one of the candidate constraints, or ignore the warning and attempt to perform the task assignment without changing any of the constraints.  

%\begin{figure*}[ht!]
%\centering
%\includegraphics[width=\textwidth]{gfx/overcooked.png} % Reduce the figure size so that it is slightly narrower than the column.
%\caption{An example of the Overcooked environment. For an evaluation of our system, we would extend layouts used in previous papers to include more tasks and (potentially) more than two agents. Figure adapted from \citet{endtoendWeeklyDexterous}}
%\label{fig:overcooked}
%\end{figure*}

\begin{figure}[t]
\centering
\includegraphics[width=\columnwidth]{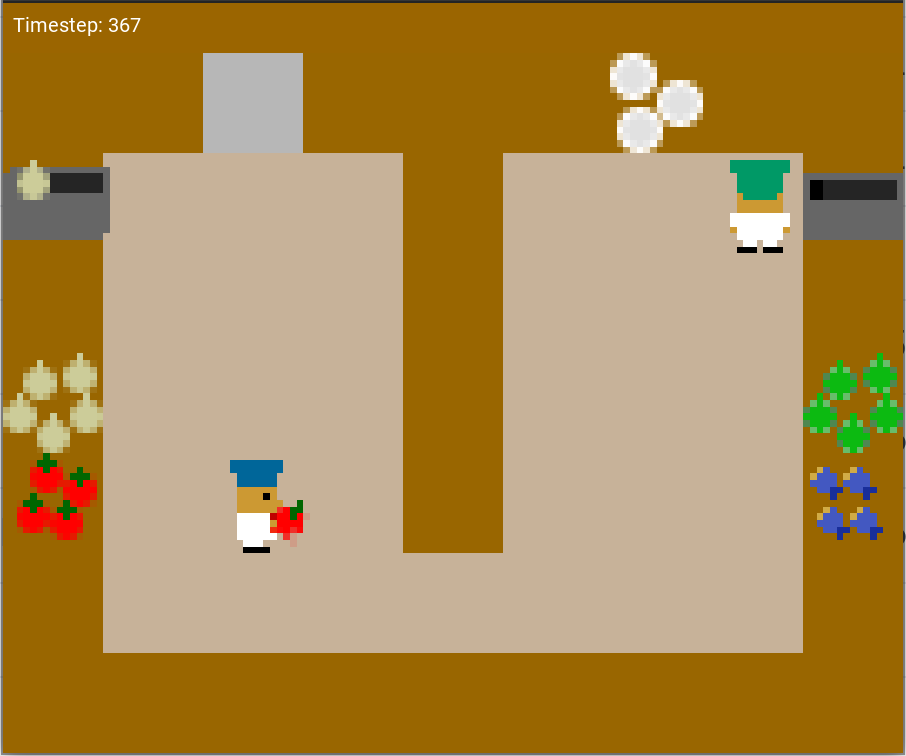} % Reduce the figure size so that it is slightly narrower than the column. Don't use precise values for figure width.This setup will avoid overfull boxes.
\caption{An example of the Overcooked environment. For an evaluation of our system, we would extend layouts used in previous papers to include more tasks and (potentially) more than two agents.}
\label{fig:overcooked}
\end{figure}

If the semantic check passes, the consistency monitor enters phase two. In this phase, our system will attempt to find a solution to the MILP using the constraints provided. If a solution cannot be found, the consistency monitor will perform a \textit{relaxation check}, wherein our system will attempt to solve the MILP by relaxing constraints in $C_{user}$ using, for example, a branch and bound relaxation approach \cite{clausen1999branch}. While this may produce a solution, the solution may violate the user's stated expectations for the task assignment. Thus, the consistency monitor sends the relaxed constraints to the language interface, wherein it will attempt to produce  natural language explanations of the new constraints using its LLM. Much like in the first phase, the user will be prompted to accept, reject, or modify the proposed constraints.

The third and final phase is initiated when a relaxed solution cannot be found. In this phase, the consistency monitor will perform an \textit{ablation check}, wherein it will attempt to solve this MILP by iteratively removing constraints from $C_{user}$. Removed constraints are then sent to language interface, wherein a natural language explanation is generated, prompting the user to try and reformulate their commands, if they desire.

%Additionally, when prompted by the language interface (see above), the consistency monitor can perform \textit{counterfactual assignments}. A counterfactual assignment is initiated when the user issues a query to the language interface like "why wasn't agent $a$ assigned to task $j$?" In this case, the consistency monitor would attempt to make this assignment, and observe which constraints, if any, are violated. The violated constraints are then sent to the language interface, and a suitable answer to the user's query is generated.

\section{Evaluation}
We plan to run a user-study in order to evaluate our approach. In this section we describe our planned experimental setup and outline  questions we wish to be able to answer.

\subsection{Environment and Setup}

Our experiments will involve naive human users interacting with and overseeing a task-assignment system during a simulated multi-agent task. A good candidate for this task is the Overcooked environment (\citealp{carroll2019utility}; Fig. \ref{fig:overcooked}), which has been gainfully employed across machine learning and human-AI teaming research (e.g. \citealp{hu2021off, aroca2023hierarchical, pearce2023imitating}).
The Overcooked environment is a gridworld based on the popular video game \textit{Overcooked}, in which agents must “cook” various dishes by performing actions like gathering ingredients and plating dishes throughout the shared environment. The shared reward for the agents is based on the number of dishes they are collectively able to prepare within the time limit. This domain is ideal for our purposes as it rewards teamwork and coordination among agents, though is simple enough for new users to quickly understand and follow. At the same time, effectively coordinating agents can be a challenge depending on the arrangement of the environment. 

\subsection{Experiments}
With our experiments we wish to evaluate our approach along subjective and objective criteria. Our research aims can be summarized by the following three questions:
\begin{enumerate}
    \item How  accurately can our LLM-based language interface transform natural speech into valid symbolic constraints, and vice-versa?
    \item How effectively can users interact with our system to generate productive and viable assignments?
    \item How well do  users' expectations for the system's behavior align with its actual behavior, and how does this impact user perceptions of the system overall? 
\end{enumerate}

In order to answer these questions, we employ a between-subjects user study. In our two experimental conditions, participants will be tasked with overseeing a game of Overcooked by providing supervisory constraints to teams of AI-controlled agents, with the goal of maximizing the agents' score across all rounds. In the first condition, users will provide these constraints using the language interface described above. In the second condition, users will be interacting with a system inspired by \cite{petersen2013fast}, where desired constrained are selected and applied using a drop-down GUI interface. Our control condition will have the task assignment and scheduling system running autonomously, without user oversight.
%While participants' speech will not be restricted, they will be informed of the scope of commands that the system can effectively understand and integrate. In the experimental condition, subjects will be interacting with the system described in this paper.

Answering question 1. will require us collect natural language commands from participants and their ground-truth constraint counterpart, and comparing them against the output to our model. Empirically quantifying the accuracy of the model in this regard is critical, as the ability to effectively translate between user speech and symbolic constraints  is central to the utility of our approach.  Our primary empirical result will come from an exploration of question 2. We intend to compare participants across our experimental and control conditions in terms how well the agents perform in the Overcooked environment. This will allow us to quantify the performance impact a human-in-the-loop has on the assignment process as a function of interaction paradigm. Question 3 will be explored via post-experiment questionnaire. This questionnaire will enable us to interrogate the potential trade-offs between performance and interactivity.

% N.B. The original paper goes into a lot of detail and has plenty of figures: https://proceedings.neurips.cc/paper/2019/hash/f5b1b89d98b7286673128a5fb112cb9a-Abstract.html
\section{Conclusion and Future Work}

In this paper we outline ongoing work towards a flexible natural language interface for a multi-agent task-assignment system.  Our proposed system  enables users to modify and inspect the constraints on a MILP via spoken dialogue, granting them control and insight over the task assignment process. We provide a high level overview of the system architecture, and outline a planned user study for evaluating our approach.

As this work is in its preliminary stages, there are still a number of open questions and challenges that need to be resolved. For instance, the complete set of symbols that comprise the MILP and its constraints, and thus the scope of constraints the users can apply, has yet to be fully determined. Similarly maximizing the accuracy our language interface for producing constraints, while minimizing LLM ``hallucinations" \cite{dinan2018wizard} remains an ongoing challenge. Nevertheless, when our system is complete, we believe it will bridge the gap between non-technical users and task assignment systems, and foster more effective human-AI partnerships.   

\section{Acknowledgements}

This work was supported by the Army Research Laboratory under
Grant W911NF-21-2-02905 and the Office of Naval Research under Grant N00014-22-1-2482. %and W911NF-21-2-0126 and by the Office of Naval Research under Grant N00014-22-1-2482.

\bibliography{aaai23}

\end{document}